\documentclass{article}

\usepackage{arxiv}
\usepackage{textcomp}
\usepackage{cite}
\usepackage{amsmath,amssymb,amsfonts}
\usepackage{algorithm}
\usepackage{algorithmic}
\usepackage{graphicx}
\usepackage[utf8]{inputenc} 
\usepackage[T1]{fontenc}    
\usepackage{hyperref}       
\usepackage{url}            
\usepackage{booktabs}       
\usepackage{amsfonts}       
\usepackage{nicefrac}       
\usepackage{microtype}      
\usepackage{lipsum}		
\usepackage{doi}

\title{Building Decision Forest via Deep Reinforcement Learning}

\author{ {Guixuan Wen} \\
	College of Computer Science\\
	Chongqing University\\
	Chongqing, China 400044  \\
	\texttt{guixuanwen@cqu.edu.cn} \\
	\And
	{Kaigui Wu} \\
	College of Computer Science\\
	Chongqing University\\
	Chongqing, China 400044 \\
	\texttt{kaiguiwu@cqu.edu.cn} \\
}

\renewcommand{\shorttitle}{Building Decision Forest via DRL}

\hypersetup{
pdftitle={A template for the arxiv style},
pdfsubject={q-bio.NC, q-bio.QM},
pdfauthor={David S.~Hippocampus, Elias D.~Striatum},
pdfkeywords={First keyword, Second keyword, More},
}

\begin{document}
\maketitle

\begin{abstract}
	Ensemble learning methods whose base classifier is a decision tree usually belong to the bagging or boosting. However, no previous work has ever built the ensemble classifier by maximizing long-term returns to the best of our knowledge. This paper proposes a decision forest building method called MA-H-SAC-DF for binary classification via deep reinforcement learning. First, the building process is modeled as a decentralized partial observable Markov decision process, and a set of cooperative agents jointly constructs all base classifiers. Second, the global state and local observations are defined based on informations of the parent node and the current location. Last, the state-of-the-art deep reinforcement method Hybrid SAC is extended to a multi-agent system under the CTDE architecture to find an optimal decision forest building policy. The experiments indicate that MA-H-SAC-DF has the same performance as random forest, Adaboost, and GBDT on balanced datasets and outperforms them on imbalanced datasets.
\end{abstract}


\section{Introduction}
\quad Decision tree is one of the classical machine learning algorithms, which goal is to generate a tree structure used to represent a set of test rules by continuously dividing the feature space. Unlike black box models that provide only predictions, decision tree models are not only simple and easy to use, but also have natural interpretability. Users are able to obtain more detailed information about the decision process from decision tree models, which is why it is widely used in industries such as medical aid diagnosis\cite{b1, b2}, financial risk control\cite{b3, b4}, and marketing\cite{b5, b6}.

However, individual decision tree model is very prone to overfitting and often requires additional pruning to trade-off accuracy and robustness. Ensemble learning\cite{b7} is one of the common means to improve the robustness and accuracy of decision tree models. For example, in the random forest\cite{b8}, users sample multiple subsets from the dataset in parallel for training different decision tree models and eventually make decisions by voting. In addition, each tree in a random forest is fully grown without pruning. Random forests tend to perform very well, especially for those datasets that contain a large number of attributes.

Another important issue is that the data imbalance also poses a challenge for decision tree\cite{b9}. Data imbalance, also known as data skew, is one of the difficulties that machine learning algorithms must address in real-world applications. For example, in disease diagnosis, the number of healthy individuals is often 1000 or even 10,000 times greater than the number of diseased individuals, and the importance of accurately predicting diseased individuals is much higher than that of accurately predicting healthy individuals. Some classical decision tree algorithms are proposed without considering the distribution of the data, and although these algorithms can achieve good performance when dealing with balanced data, they will fail in the face of imbalanced data because the algorithm is biased towards the majority class. The reason for this phenomenon is that the split criteria introduces the prior probability of the sample distribution in the process of calculating the node impurity.

This paper proposes a deep reinforcement learning(DRL) based decision forest induction method called MA-H-SAC-DF for binary classification with the maximization of a long-term return, where a set of cooperative agents jointly constructs all base classifiers. This idea builds on our previous work\cite{b10}, which proposed to represent the decision tree induction by Markov decision process(MDP) and solve the optimal induction strategy by deep reinforcement learning algorithm with hybrid action space. In MA-H-SAC-DF, the decision forest building process is considered as a multi-step game that can be modeled as a decentralized partial observable Markov decision process(Dec-POMDP). Besides, the state space during decision forest induction is redefined according to informations of  parent node and current position. Still, the action space and reward function are consistent with our previous work. Finally, we extend the Hybrid SAC\cite{b11} to the multi-agent system based on the centralized training decentralized execution(CTDE) architecture\cite{b12} and propose MA-H-SAC, which is applied to solve the optimal decision forest induction policy. Experimental results show that MA-H-SAC-DF not only has the same performance as random forest, Adaboost and GBDT on balanced datasets, but also outperforms them on unbalanced datasets.

The rest of this paper is organized as follows. The second section firstly introduces the ensemble learning related to decision tree, and then summarizes the Hybrid SAC algorithm and the existing multi-agent method architecture. The details of MA-H-SAC and MA-HSAC-DF will be described in the third section. The fourth section shows the experimental results and evaluates the performance of MA-H-SAC-DF compared with other methods. Conclusions and future work will be discussed in section five.

\section{Reated work}
\subsection{Decision tree based ensemble learning}
\quad The basic idea of ensemble learning is to combine multiple week classifiers to form a strong classifier. The Hoffding inequality\cite{b13} proves that when the accuracy of the base classifier is more than 50\%, the error rate of the ensemble classifier decreases exponentially with the increase of the number of base classifiers and finally approaches 0. Decision tree is a common base classifier in ensemble learning which can be classified into bagging, boosting, and stacking according to different ensemble strategies.

One of the earliest ensemble algorithms is bagging\cite{b15}, where each base classifier is trained on a subset of the initial training set, and new samples are usually predicted by voting. Random forest, whose base classifier is a decision tree, is a variant of bagging. Like bagging, random forest uses bootstrap sampling and aggregates the predictions of all base classifiers in an unweighted manner for the final vote. However, the induction of the base classifier in the random forest is different from that of a single decision tree. Firstly, $m$ attribute subsets are randomly selected from $M$ attributes, and then base classifiers are trained on the sampled data subsets based on attribute subsets. The best value for $m$ is always $\sqrt{M}$. In addition, every tree in the random forest is fully grown without pruning. Random forests tend to perform very well, especially for data sets that contain many attributes.

Unlike the bagging, the base classifiers are generated sequentially in the Boosting\cite{b15}. Firstly a base classifier is trained from the initial training set, and then the next base classifier is trained based on the performance of the previous base classifier. This process is repeated until the number of base classifiers reaches a pre-specified. The final set of base classifiers is weighted and combined. Gradient Boosting Decision Tree (GBDT)\cite{b14} is a classical algorithm in the boosting. For the classification problem, given a data set $D=\left\{ x_i,y_i \right\} ,i=1,..,n$, the loss function of a single sample can be expressed in terms of cross-entropy 
\begin{equation}
L\left( x_i,y_i \right) =-y_i\log \hat{y}_i-\left( 1-y_i \right) \log \left( 1-\hat{y}_i \right) 
\end{equation}
where $\hat{y}_i$ denotes the prediction probability. Assuming that the $m$th base classifier is $h_m (x)$, the ensemble classifier obtained in the first $k$ iterations is 
\begin{equation}
F\left( x \right) =\sum_{m=0}^k{h_m\left( x \right)}
\end{equation}
According to the log odds function $\hat{y}=\frac{1}{1+e^{-z}}$, substitution gives
\begin{equation}
L\left( x_i,y_i|F\left( x \right) \right) = y_ilog\left( 1+e^{-F\left( x_i \right)} \right) + \left( 1-y_i \right) \left[ F\left( x_i \right) +log\left( 1+e^{-F\left( x_i \right)} \right) \right]
\end{equation} 
It is easy to obtain that the negative gradient of the loss function L for the current ensemble classifier is
\begin{equation}
\left. \frac{-\partial L}{\partial F\left( x \right)} \right|\left( x_i,y_i \right) =y_i-\frac{1}{1+e^{-F\left( x \right)}}=y_i-\hat{y}_i
\end{equation} 
Then the sample $\left\{ x_i,y_i-\hat{y}_i \right\} _{i=1}^{n}$ is used as the training sample for the $m+1$th base classifier. This is repeated until the number of base classifiers reaches a pre-specified value.

Stacking\cite{b15} is different from bagging and boosting in two main points. First, stacking considers heterogeneous base classifiers, while bagging and boosting consider homogeneous base classifiers. Second, stacking uses meta model to combine base classifiers. For example, KNN, logistic regression, and SVM can be selected as the base classifier and the neural network as the meta model. The neural network takes the output of the base classifier as input and produces the final prediction result.  

\subsection{Hybrid SAC}
\begin{figure}[h] 
\centering 
\includegraphics[width=0.5\columnwidth]{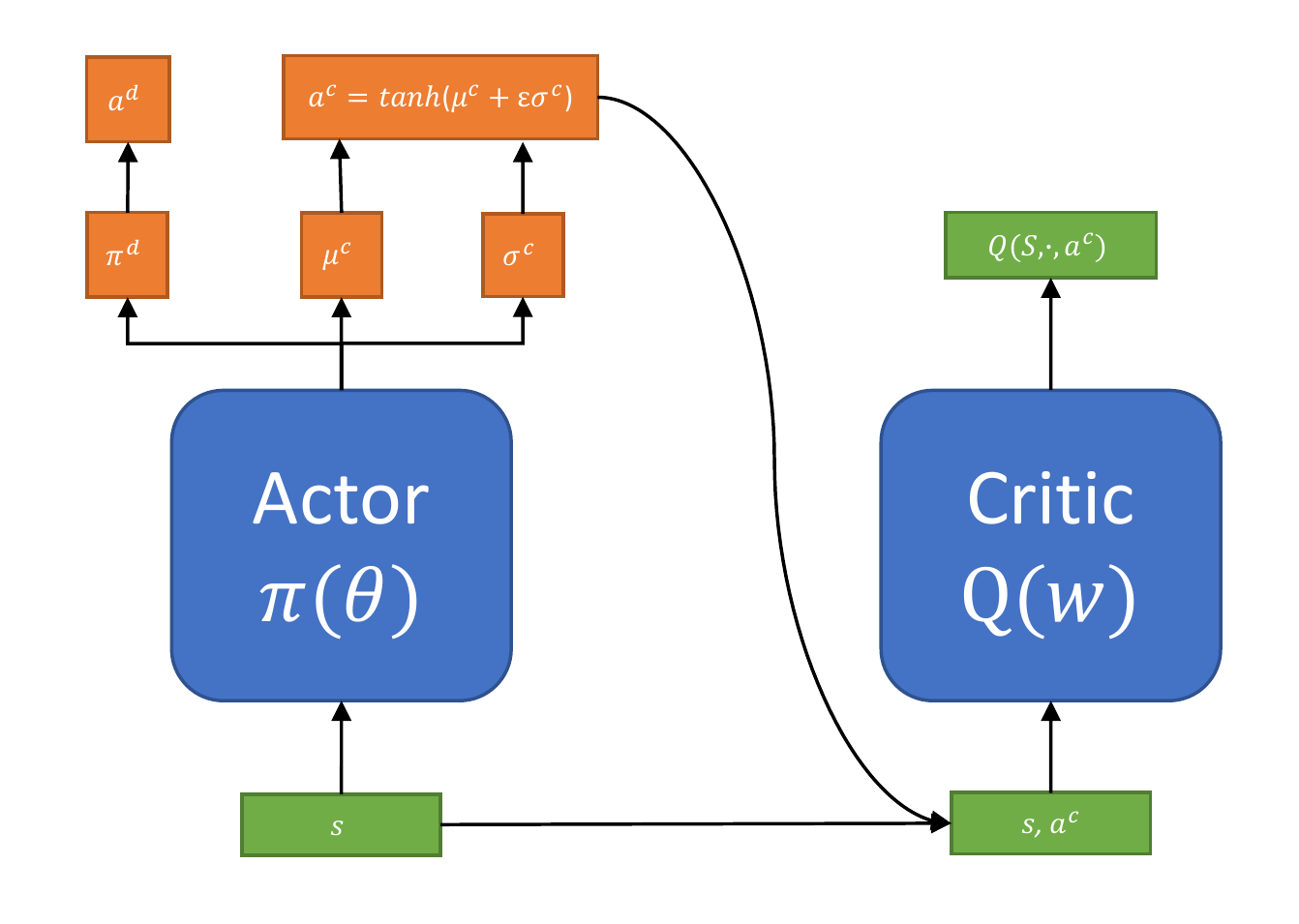} 
\caption{The framework of hybrid sac} 
\label{fig1}
\end{figure}

\quad Many deep reinforcement learning methods only consider the discrete or continuous action, but hybrid action space is also essential. Hybrid SAC\cite{b11}, an extension of SAC, can directly handle hybrid action space without any approximation or relaxation. Assuming that there are $K$ discrete actions and each action corresponds to an $m$-dimensional parameter $x_k$, the hybrid action space can be expressed as
\begin{equation}
\mathcal{A}=\left\{ \left( k,x_k \right) |x_k\in \mathcal{X}_k\subset R^m\ for\ all\ k\in \left[ K \right] \right\} .
\end{equation}
As depicted in the FIGURE \ref{fig1}, Hybrid SAC follows the actor-critic architecture, where the actor network takes the state as input and produces both a discrete distribution $\pi^d$ as well as the mean $\mu^c$ and standard deviation $\sigma ^c$ of the continuous parameters, and the critic network takes state combined with parameters actions as input, then produces $q$ values for all discrete actions.

\subsection{Multi-agent architecture}
\quad In multi-agent system (MAS), different agents interact in a common environment and make decisions autonomously to achieve their own goals or a common goal. According to different tasks, the relationship between agents can be divided into cooperation, competition and mix. Compared with the single agent system, the application of reinforcement learning in MAS will face the following challenges: the instability of the environment, the limitation of the agent to obtain information, the consistency of individual goals and scalability\cite{b16}.

Distributed training decentralized execution (DTDE) is a multi-agent deep reinforcement learning architecture proposed earlier. In DTDE, the training stage of each agent is independent, and there is no information sharing between agents. Besides, each agent makes independent decisions during the execution stage based on its local observations. The disadvantage of the DTDE architecture is that the environment is unstable for any agent\cite{b17}. Because while an agent is making decisions, other agents are also taking actions, which leads to changes in the state of the environment related to all agents' joint actions.

Centralized training centralized Execution (CTCE),  which is the opposite of DTDE, assumes that all agents can communicate with each other. Therefore, it can learn a joint policy for all agents by directly applying the single agent method to MAS\cite{b17}. Although methods based on CTCE architecture can improve scalability to some extent by sharing parameters, they still face the dimension disaster of exponential growth of joint action space as the number of agents increases.

Centralized training decentralized execution (CTDE) is the most popular multi-agent training architecture. Each agent must train with the global information and make independent decisions based on local observations during execution. Jakob Foerster et al.\cite{b18} first applied the training method of centralized critic distributed actor based on the actor-critic architecture and proposed counterfactual multi-agent policy gradients (COMA) algorithm. In addition, a series of landmark multi-agent deep reinforcement learning methods such as MADDPG\cite{b12}, VDN\cite{b19}, QMIX\cite{b20}, and QTRAN\cite{b21} is also built based on the CTDE architecture. Therefore, the MA-H-SAC proposed in this paper is also on CTDE.

\section{Methods}
\quad Ensemble learning is one of the standard methods to improve the generalization and robustness of algorithms. So far, the ensemble algorithms whose base classifier is decision trees usually belong to the bagging or boosting. There has been no similar work to build an ensemble classifier to maximize a long-term return. This paper first comes up with a multi-agent deep reinforcement learning method called MA-H-SAC for hybrid action space. It then proposes a new decision forest induction method based on MA-H-SAC called MA-H-SAC-DF, where a group of cooperative agents constructs all base classifiers.
\subsection{MA-H-SAC}
\quad Usually, a multi-agent task can be described as the decentralized partially observable Markov decision process (Dec-POMDP), which can be represented by tuples $(S,A,P,R,O,n,\gamma)$, where $s\in S$ represents the global state of the environment. Considering that the agent is partially visible to the environment, at each time-step $t$, the agent $i \in I \equiv \left\{1,2,...,n\right\}$ obtains its observation $o_i \in O_i$ and selects an action $a_i \in A_i$, which forms a joint action $a \in A$. When the joint action $a \in A$ is performed, the environment transitions according to the state transition function $P(s^\prime\left.  \right|s,a):s\times a \times s \rightarrow [0,1]$. All agents share the same reward function, $R(s,a): s \times a \rightarrow R$ and discount rate $\gamma$.

\begin{figure}[htb] 
\centering 
\includegraphics[width=0.5\columnwidth]{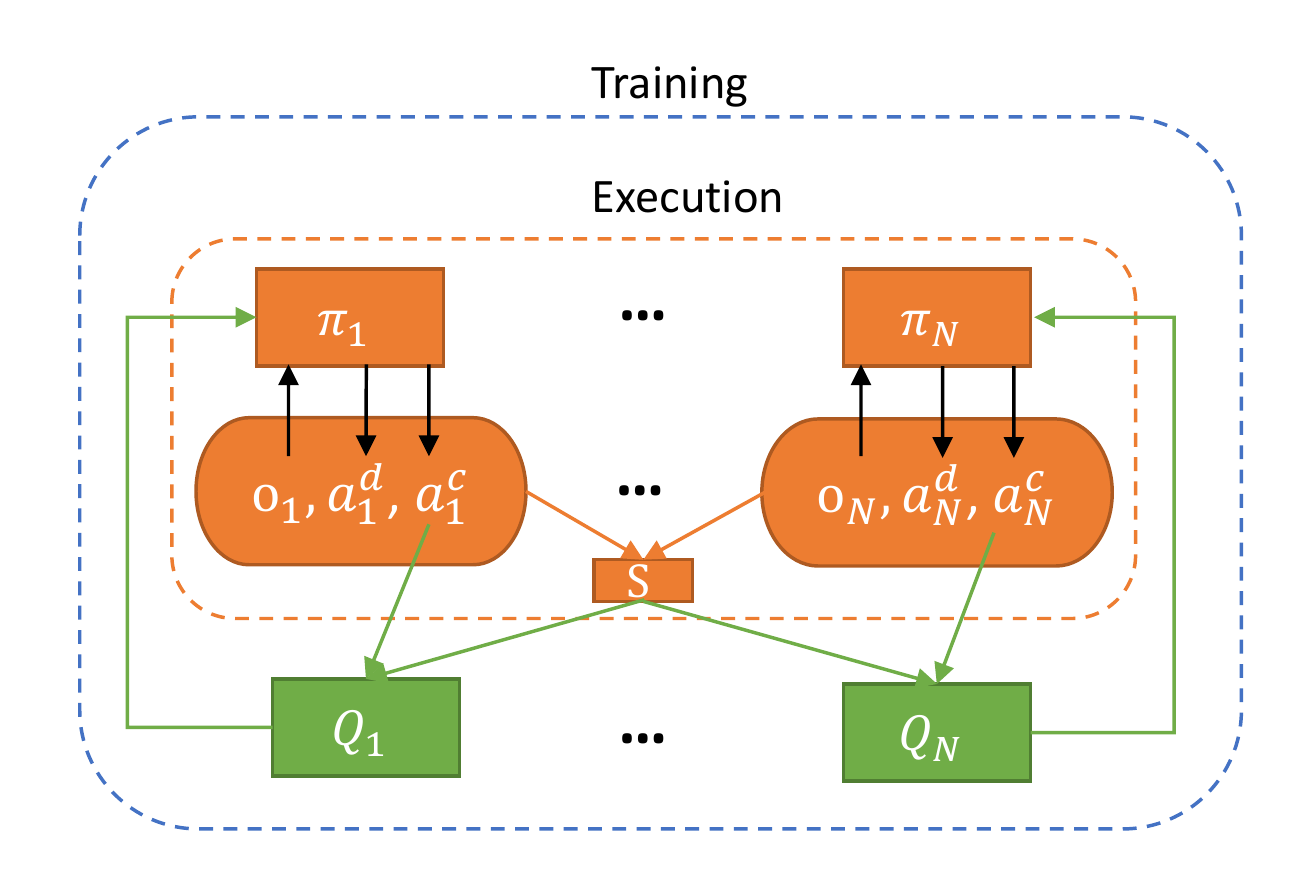} 
\caption{The framework of MA-H-SAC} 
\label{fig2}
\end{figure}

MA-H-SAC  is an extension of Hybrid SAC to the MAS under the architecture of CTDE. The global state is utilized in the training stage to ensure that the environment is stable. Each agent only needs to use the local observation to make decisions during execution. Suppose $n$ agents with the actor network $\pi=(\pi_1,\pi_2...,\pi_n)$ and critic network $Q=(Q_1, Q_2..., Q_n)$. As shown in the FIGURE \ref{fig2}, like MADDPG, each agent in MA-H-SAC uses an independent critic network $Q_i(s,\cdot,a_{i}^{c};w_i)$ that takes the global state $s$ and continuous actions $a_{i}^{c}$ as input and produces the $q$ values of all discrete actions during the centralized training stage. In the decentralized execution stage, the actor network can simultaneously output the probability distribution $\pi_{i}^{d}(\cdot \left.  \right|o_i)$ of discrete actions as well as the mean $\mu_{i}^{c}$ and standard deviation $\theta_{i}^{c}$ of all continuous actions only by inputting local observations. Then the agent obtains specific discrete actions by sampling from the probability distribution, i.e. $a_{i}^{d} \backsim \pi_{i}^{d}$, and computes the final continuous parameters by the resample trick $a_{i}^{c}=tanh(\mu_{i}^{c} + \epsilon \theta_{i}^{c})$, where $\epsilon$ is the gaussian noise.

According to the CTDE architecture and Hybrid SAC algorithm, the loss function of the $i$-th agent for updating actor network $\pi_i$ is 
\begin{equation}
\label{eq6}
J_{\pi _i}\left( w_i \right)=\mathbb{E}_{\left( s,o \right) \thicksim B}\left[ \mathbb{E}_{\left( a_{i}^{d},a_{i}^{c} \right) \thicksim \pi _i}\left[ \mathbb{H}^d+\mathbb{H}^c
-2\mathbb{Q} \right] \right]
\end{equation}
where 
\begin{eqnarray}
\label{eq7}
\mathbb{Q} =& Q_i\left(s,a_{i}^{d},a_{i}^{c};w_i\right)\\
\mathbb{H}^d =& \alpha ^d\log \left[ \pi _i\left( \left. a_{i}^{d} \right|o_i;\theta _i \right) \right] \\
\mathbb{H}^c =& \alpha ^c\pi _i\left( \left. a_{i}^{d} \right|o_i;\theta _i \right) \log \left[ \pi _i\left( \left. a_{i}^{c} \right|o_i,a_{i}^{d};\theta _i \right) \right] 
\end{eqnarray}
$\alpha^d$ and $\alpha^c$ are temperature parameters, and the replay buffer $B$ is used to store the agent's historical experience $(S, O, A, R, S ^\prime, O ^\prime)$. The parameter $w_i$ of the critic network $Q_i$ can be updated by minimizing the Bellman error
\begin{equation}
\label{eq8}
J_{Q_i}\left( w_i \right) =\mathbb{E}_{\left( s,o \right) \thicksim B}\left[ \frac{1}{2}\left( \mathbb{E}_{\left( a_{i}^{d},a_{i}^{c} \right) \thicksim \pi _i}\left[ \mathbb{Q} \right] -y \right) ^2 \right]
\end{equation}
where
\begin{equation}
\label{eq9}
y=r+\gamma \mathbb{E}_{\left( a_{i}^{'d},a_{i}^{'c} \right) \thicksim \pi _i}\left[ Q_i\left( s^{'},a_{i}^{'d},a_{i}^{'c};w_{i}^{'} \right) -\mathbb{H}^{'d}-\mathbb{H}^{'c} \right] 
\end{equation}
$w_{i}^{'}$ is the target critic network parameters. It is worth noting that, to make the training more stable, MA-H-SAC refers to the suggestions in TD3\cite{b22} that uses double critic network to avoid the overestimation problem of value function during training. The details of MA-H-SAC are shown in Algorithm \ref{Algorithm 1}.

\begin{algorithm}
\caption{MA-H-SAC}
\label{Algorithm 1}
\begin{algorithmic}[1]
\REQUIRE ~~\\ 
\STATE learning rate $l_{d}, l^{c}$, max episode $M$, max timestep $T$, Gaussian distribution $\mathcal{N}$, temperature factor $\alpha^{d},\alpha^{c}$, number of agents $n$
\ENSURE ~~\\ 
\STATE Initialize actor network $\pi=(\pi_1(\theta_1),\pi_2(\theta_2)...,\pi_n(\theta_n))$, critic network $Q^{j}=(Q_1^{j}(w_1^{j}),Q_2^{j}(w_2^{j})...,Q_n^{j}(w_n^{j})), j \in \left\{1,2 \right\}$
\STATE Initialize target actor network and critic network
\FOR{$episode = 0$ to $M-1$}
\STATE obtain initial global state $s_0$ and local observations $o_{0}=(o_{0,1},o_{0,2},...,o_{0,n})$
\FOR{$t = 0$ to $T-1$}
\FOR{$i = 0$ to $n-1$}
\STATE compute $(\pi_{t,i}^{d},\mu_{t,i}^{c},\sigma_{t,i}^{c}) = \pi_i(o_{t,i};\theta_{i})$
\STATE select action $a_{t,i}=(a_{t,i}^{d}, a_{t,i}^{c})$, where
$$
a_{t,i}^{d} \backsim \pi_{t,i}^{d}, a_{t,i}^{c} \gets tanh(\mu_{t,i}^{c}, \sigma_{t,i}^{c})
$$
\ENDFOR
\STATE takes the joint action $a_t=(a_{t,1},a_{t,2},...,a_{t,n})$,  obtain reward $r_t$, next global state $s_{t+1}$ and next local observations $o_{t+1}$
\STATE save tuple $(s_t, o_t, a_t, r_t, s_{t+1}, o_{t+1})$ to replay buffer $B$
\FOR{$i = 0$ to $n-1$}
\STATE get a batch of samples $(s, o, a, r, s^{\prime}, o^{\prime})$ from $B$
\STATE compute the target $y$ according to Equation \eqref{eq9}
\STATE compute the loss and update critic network $Q_{i}^{j}(w_{i}^{j})$ according to Equation \eqref{eq8}
\STATE compute the loss and update actor network $\pi_{i}(\theta_{i})$ according to Equation \eqref{eq6}
\STATE update the target network
$$
w_{i}^{\prime, j}=\tau w_{i}^{j} + (1-\tau)w_{i}^{\prime, j}
$$
\ENDFOR
\ENDFOR
\ENDFOR
\end{algorithmic}
\end{algorithm}
\subsection{MA-H-SAC-DT}
\begin{figure}[htb] 
\centering 
\includegraphics[width=0.5\columnwidth]{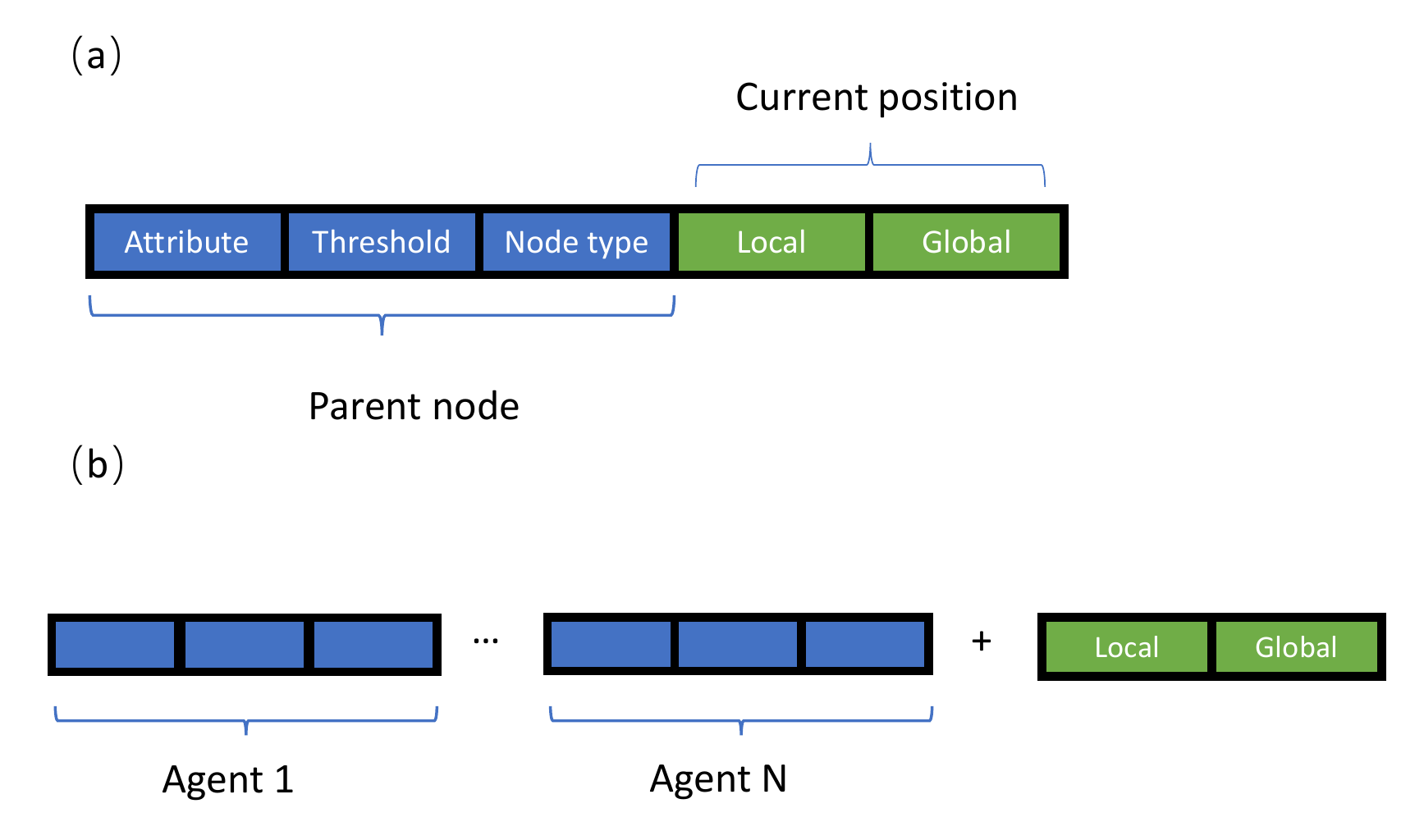} 
\caption{Representation of partial observation and global state of decision forest} 
\label{fig3}
\end{figure}

\quad MA-H-SAC-DF needs to transform the induction process of decision forest into DEC-POMDP first. As shown in FIGURE \ref{fig3} (a), it is assumed that $n$ agents respectively make $n$ trees. When the agent generates a node, the local observation information includes the parent node's attribute, threshold value, and node type. In addition, local observation also consists of the node's local position and global position information. The local position means that the current node is a left child or a right child. The global position refers to the current node's position when traversing the whole decision tree in a level order traversal way. All the information is represented by One-Hot encoding and spliced into the one-dimensional vector to form the local observation vector $o_i$ of agent $i$. MA-H-SAC adopts the CTDE architecture, and the global state of the environment is required as input of the critic network during centralized training. A simple method is to all of the agent's local observations in terms of joining together the global state, which is easy to produce redundant information. Because all the agents that generate nodes by means of level order traversal, at the same time, the node's local and global position are the same, it can only keep one part of this information. The simplified global state is shown in FIGURE \ref{fig3} (b).

The action space and reward function are defined as same as when building a single decision tree model which is based on our previous work. We consider the continues attributes in binary classification. Thus the attributes can be represented as discrete action and the threshold values are the corresponding continues action-parameter $x_k\in X_k\subseteq R$. Each node in DT corresponds with a action $a_k=\left( k,\ x_k \right)$. 
\begin{equation}A_i=\underset{k\in \left[ K \right]}{\cup}\left\{ a_k=\left( k,x_k \right) |x_k\subseteq X_k \right\} 
\label{eq10}\end{equation}

Intuitively we can combine the tree structures generated by agents into ensemble classifiers and classify on the training
set at step $t$. Next, the predicted results $\hat{Y}_t$ and the truth $Y_t$ are used to calculate a score $sc_t$ based on arbitrary evaluation metrics, such as accuracy and G-Mean. Finally the reward $r_t$ is easily obtained according to equation \eqref{eq11}.\begin{figure*}[htb] 
\centering 
\includegraphics[width=0.8\columnwidth]{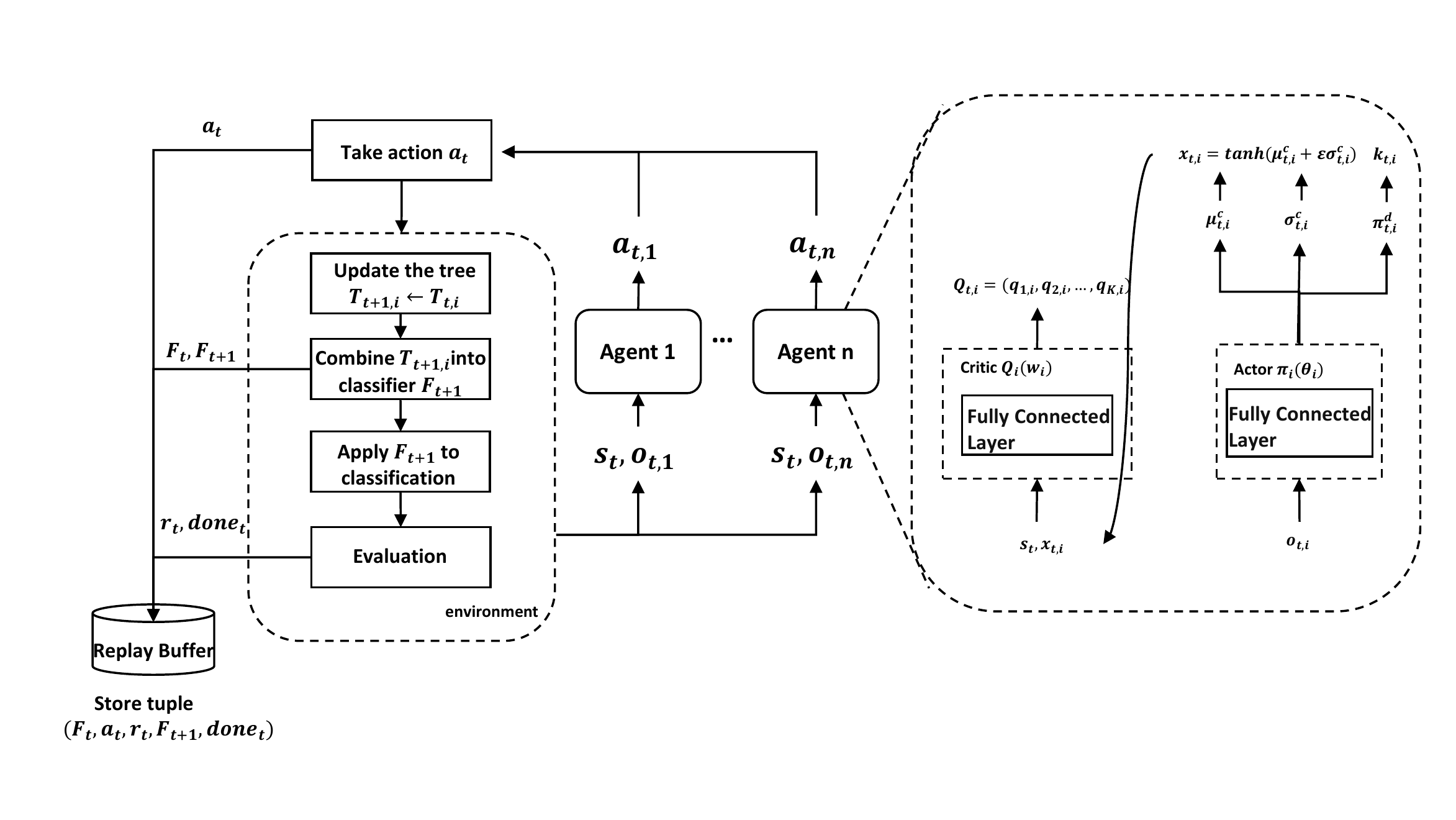} 
\caption{The Framework of MA-H-SAC-DT} 
\label{fig4}
\end{figure*}That is to say, the positive $r_t$ means action $a_t$ improves the performance of the decision tree
while negative reward $r_t$ decreases it. Note that the $sc_0$ is usually set to 0 or 0.5. In other words, if the initial $sc_0$ is set to zero, the total reward $R$ is equivalent to the evaluation score of the final classifier. Similarly, setting $sc_0$ at 0.5 is equivalent to adding a baseline.
\begin{equation}r_t=sc_t-sc_{t-1}
\label{eq11}\end{equation}

The framework of MA-H-SAC-DF is shown in FIGURE \ref{fig4}. Each agent is responsible for the generation of a corresponding base classifier. At time $t$, the agent $i$ selects attribute $k_{t, i}$ and its corresponding threshold value $x_{t,k,i}$ according to actor network to form action $a_{t,i}=(k_{t, i},k_{t,k, i})$, and all agent actions form joint action $a_t$. The environment generates node $node_{t, i}$  for each base classifier according to the attribute and partition threshold in the joint action performed by the agent. Next, the environment combines all base classifiers into ensemble classifiers in an unweighted way for classification evaluation in dataset $D$ and gets the reward $r_t$. Finally, the agent randomly samples a group of samples from the experience pool to update the network parameters. It should be noted that MA-H-SAC-DF adopts a delayed update strategy to update the actor network of an agent, which is similar to the MATD3 method.
\section{Experiment}
\subsection{Datasets}
\quad In this section, random forest, GBDT, and Adaboost are used as comparison baselines to verify the performance of MA-H-SAC-DT in both balanced data and imbalanced data. \begin{figure}[htb] 
    
  \begin{minipage}[b]{\textwidth} 
    \centering
    \caption{Description of the balanced datasets. \#I, \#F denote the number of instances, attributes respectively}\label{tab1}
    \begin{tabular}{|p{0.3\textwidth}|p{0.1\textwidth}|p{0.1\textwidth}|}
\hline
	Data Sets& 
	\#I&
	\#F\\
	\hline
	appendicitis& 
	106& 
	7\\
	bupa& 
	345& 
	6 \\
	coil2000& 
	9822& 
	85 \\
	heart& 
	220& 
	13 \\
	magic& 
	19020& 
	10 \\
	pima & 
	768& 
	8 \\
	sonar1& 
	208& 
	60 \\
	spectfheart& 
	267& 
	44 \\ 
	breast-cancer& 
	683& 
	10 \\
	bands& 
	365& 
	19 \\
	australian& 
	690& 
	14 \\
	mammographi& 
	830& 
	5 \\
	saheart& 
	462& 
	9 \\
	liver-disorders& 
	345& 
	5 \\
	diabetes& 
	759& 
	8 \\
	\hline
\end{tabular}
  \end{minipage} 
  \begin{minipage}[b]{\textwidth} 
    \centering
    \caption{Description of the imbalanced datasets. \#I, \#F denote the number of instances, attributes respectively}\label{tab2}
    \begin{tabular}{|p{0.3\textwidth}|p{0.1\textwidth}|p{0.1\textwidth}|p{0.1\textwidth}|}
    \hline
    	Data Sets& 
    	\#I&
    	\#F&
    	IR \\
    	\hline
    	ecoli-0-1-vs-2-3-5& 
    	244& 
    	7& 
    	9.17 \\
    	ecoli-0-1-4-6-vs-5& 
    	187& 
    	6& 
    	13.0 \\
    	ecoli-0-1-4-7-vs-2-3-5-6& 
    	336& 
    	7& 
    	10.59 \\
    	ecoli-0-6-7-vs-5& 
    	220& 
    	6& 
    	10.0 \\
    	ecoli2& 
    	336& 
    	7& 
    	5.46 \\
    	haberman& 
    	306& 
    	3& 
    	2.78 \\
    	new-thyroid1& 
    	215& 
    	5& 
    	5.14 \\
    	new-thyroid2& 
    	215& 
    	5& 
    	5.14 \\
    	vehicle3& 
    	846& 
    	18& 
    	3.0 \\
    	winequality-red-4& 
    	1599& 
    	11& 
    	29.17 \\
    	wisconsin& 
    	683& 
    	9& 
    	1.86 \\
    	yeast-0-2-5-6-vs-3-7-8-9& 
    	1004& 
    	8& 
    	9.14 \\
    	yeast1& 
    	1484& 
    	8& 
    	2.46 \\
    	glass0& 
    	214& 
    	9& 
    	2.06 \\
    	glass6& 
    	214& 
    	9& 
    	6.38 \\
    	pima& 
    	768& 
    	8& 
    	1.87 \\
    	africa recession& 
    	486& 
    	53& 
    	11.9 \\
    	insurance& 
    	382154& 
    	10& 
    	5.1 \\
    	\hline
    \end{tabular} 
  \end{minipage}
\end{figure}15 balanced datasets and 18 imbalanced datasets from reality are described in TABLE \ref{tab1} and TABLE \ref{tab2}, which contains the number of instances and attributes. Besides, the imbalance ratio (IR) that measures the degree of imbalance between the classes of majority and minority is also provided. These data sets are collected from three well-known public sources called UCI Machine Learning Repository, and KEEL Imbalanced Data Sets.
\subsection{Evaluation Metrics}
\quad We use the accuracy  to evaluate the model's performance on balanced data sets. However, it is unreasonable to take the accuracy as the metric to evaluate classifier performance in imbalanced classification. At present, the commonly used imbalanced classification evaluation metrics include G-Mean and the area under the ROC curve (AUC) that are applied in our experiments.

To make the experiment more convincing, we not only take 10-fold cross-validation but also conduct a Friedman test and Nemenyi test. The null hypothesis of the Friedman test is that all the methods are equivalent. Precisely, assuming there are $k$ methods, $N$ data sets, and average rank $r_i$ corresponding to each method $m_i$, we should compute two essential statistics $\chi _{F}^{2}$ and $F_F$ that calculated as \eqref{eq12} and \eqref{eq13}, then compare the value of $F_F$ with the critical value of given significance level $\alpha$. If the null hypothesis is rejected, we need to take a Nemenyi test for further comparison. For given significance level $\alpha$, the critical value $CD$ can be calculated as \eqref{eq14} on the Nemenyi test. If the average rank difference between the two methods is greater than $CD$, it is believed that the two methods have different performances.
\begin{equation}\chi _{F}^{2}=\frac{12N}{k\left( k+1 \right)}\left( \sum_{i=1}^k{r_i^{2}}-\frac{k\left( k+1 \right) ^2}{4} \right) 
\label{eq12}\end{equation}
\begin{equation}F_F=\frac{\left( N-1 \right) \chi _{F}^{2}}{N\left( k-1 \right) -\chi _{F}^{2}}
\label{eq13}\end{equation}
\begin{equation}CD=q_{\alpha}\sqrt{\frac{k\left( k+1 \right)}{6N}}
\label{eq14}\end{equation}

\subsection{Result}
 \quad The results of Friedman test(Fr.T) with the "Base" classifier are shown in the last line of tables. The "\checkmark" sign under a classifier indicates that the "Base" classifier significantly outperforms that classifier at 90\% confidence level. 
 \begin{table}[hbp]
\begin{center}
\caption{Accuracy of four methods on fifteen balanced datasets}\label{tab3}
\begin{tabular}{|p{0.3\columnwidth}|p{0.1\columnwidth}|p{0.1\columnwidth}|p{0.1\columnwidth}|p{0.1\columnwidth}|}
\hline
Datasets& 
Random Forest&
Adaboost&
GBDT&
MA-H-SAC-DT\\
\hline
appendicitis& 
0.844& 
0.833& 
0.844&
0.846\\
bupa& 
0.670& 
0.615& 
0.670&
0.583\\
coil2000& 
0.922& 
0.940& 
0.940&
0.940\\
heart& 
0.784& 
0.828& 
0.799&
0.798\\
magic& 
0.858& 
0.774& 
0.786&
0.798\\
pima& 
0.734& 
0.744& 
0.735&
0.777\\
sonar& 
0.747& 
0.750& 
0.729&
0.772\\
spectfheart& 
0.791& 
0.783& 
0.797&
0.793\\
breast-cancer& 
0.963& 
0.949& 
0.943&
0.973\\
australian& 
0.963& 
0.949& 
0.943&
0.973\\
bands& 
0.687& 
0.661& 
0.632&
0.661\\
bands& 
0.687& 
0.661& 
0.632&
0.661\\
mammographi& 
0.787& 
0.838& 
0.834&
0.833\\
saheart& 
0.648& 
0.693& 
0.688&
0.726\\
liver-disorders& 
0.580& 
0.550& 
0.588&
0.580\\
diabetess& 
0.708& 
0.729& 
0.753&
0.657\\
\hline
Avg.Accuracy&
0.772&
0.770&
0.773&
0.779\\
Avg.Rank&
2.800&
2.467&
2.600&
2.133\\
Win/Tie/Loss&
10/1/4&
9/2/4&
9/1/5&
Base\\
Fr.T&
-&
-&
-&
Base\\
\hline
\end{tabular}
\end{center}
\end{table}
 TABLE \ref{tab3} shows the performance comparison of four methods on balanced data sets according to accuracy. From the average accuracy ranking, the performance of MA-H-SAC-DF on balanced data is not different from random forest, Adaboost, and GBDT. According to equations \eqref{eq12} and \eqref{eq13}, the statistics $\mathcal{X}_F^{2}=2.120$ and $F_F=0.692$ can be calculated. In the case of 15 datasets and four methods, the critical value of Friedman test $F(3,42)=2.219$ when $\alpha=0.1$ .Because $F_F<F_{\alpha=0.1}$, the Friedman test null hypothesis is accepted, and the performance of the four methods are the same, which means that MA-H-SAC-DF has the same classification performance as random forest, Adaboost, and GBDT on balanced data.

TABLE \ref{tab4} and TABLE \ref{tab4} compare the performance differences of MA-H-SAC-DF with random forest, Adaboost, and GBDT in processing unbalanced data according to G-mean and AUC, respectively. According to the mean and average ranking of G-mean and AUC, MA-H-SAC-DF has a better classification ability. To further verify this conclusion, the Friedman test was performed on the experimental results of G-mean value and AUC value, respectively. For G-mean, $\mathcal{X}_F^{2}=39.600$ and $F_F=46.750$ can be calculated according to equations \eqref{eq12} and \eqref{eq13}. When $\alpha=0.1$, the critical value of Friedman test $F(3,51)=2.194$ when comparing the four methods on 18 datasets. Because $F_F>F_{\alpha=0.1}$, the Friedman test null hypothesis is rejected, and the performance of the four methord is not the same. For AUC, the statistics were $\mathcal{X}_F^{2}=38.467$ and $F_F=42.099>.F_{alpha = 0.1}$. We also reject the null hypothesis. According to equation \ref{eq14}, the Nemenyi test critical value $CD=0.986$, and the performance of the MA-H-SAC-DF is significantly better than the other three ensemble learning methods.
\begin{table}[hbp]
\begin{center}
\caption{G-Mean of four methods on eighteen imbalanced datasets}\label{tab4}
\begin{tabular}{|p{0.3\columnwidth}|p{0.1\columnwidth}|p{0.1\columnwidth}|p{0.1\columnwidth}|p{0.1\columnwidth}|}
\hline
Datasets& 
Random Forest&
Adaboost&
GBDT&
MA-H-SAC-DT\\
\hline
ecoli-0-1-vs-2-3-5& 
0.878& 
0.854& 
0.608&
0.938\\
ecoli-0-1-4-6-vs-5& 
0.735& 
0.750& 
0.430&
0.971\\
ecoli-0-1-4-7-vs-2-3-5-6& 
0.750& 
0.793& 
0.664&
0.891\\
ecoli-0-6-7-vs-5& 
0.772& 
0.756& 
0.656&
0.929\\
ecoli2& 
0.847& 
0.795& 
0.657&
0.886\\
haberman& 
0.475& 
0.467& 
0.060&
0.635\\
new-thyroid1& 
0.961& 
0.941& 
0.942&
0.984\\
new-thyroid2& 
0.937& 
1.000& 
0.866&
0.958\\
vehicle3& 
0.691& 
0.691& 
0.267&
0.696\\
winequality-red-4& 
0.179& 
0.000& 
0.229&
0.717\\
wisconsin4& 
0.968& 
0.973& 
0.964&
0.977\\
yeast-0-2-5-6-vs 3-7-8-9& 
0.625& 
0.435& 
0.499&
0.835\\
yeast1& 
0.619& 
0.542& 
0.000&
0.687\\
glass0& 
0.800& 
0.718& 
0.471&
0.754\\
glass6& 
0.851& 
0.894& 
0.768&
0.992\\
pima& 
0.674& 
0.683& 
0.531&
0.731\\
africa recession& 
0.383& 
0.280& 
0.275&
0.607\\
insurance& 
0.601& 
0.000& 
0.000&
0.818\\
\hline
Avg.G-Mean&
0.708&
0.643&
0.494&
0.834\\
Avg.Rank&
2.333&
2.778&
2.778&
1.111\\
Win/Tie/Loss&
17/0/1&
17/0/1&
18/0/0&
Base\\
Fr.T&
\checkmark&
\checkmark&
\checkmark&
Base\\
\hline
\end{tabular}
\end{center}
\end{table}

\begin{table}[!t]
\begin{center}
\caption{AUC of four methods on eighteen imbalanced datasets}\label{tab5}
\begin{tabular}{|p{0.3\columnwidth}|p{0.1\columnwidth}|p{0.1\columnwidth}|p{0.1\columnwidth}|p{0.1\columnwidth}|}
\hline
Datasets& 
Random Forest&
Adaboost&
GBDT&
MA-H-SAC-DT\\
\hline
ecoli-0-1-vs-2-3-5& 
0.887& 
0.861& 
0.681&
0.935\\
ecoli-0-1-4-6-vs-5& 
0.777& 
0.777& 
0.595&
0.970\\
ecoli-0-1-4-7-vs-2-3-5-6& 
0.786& 
0.820& 
0.725&
0.888\\
ecoli-0-6-7-vs-5& 
0.802& 
0.786& 
0.715&
0.926\\
ecoli2& 
0.857& 
0.812& 
0.749&
0.883\\
haberman& 
0.549& 
0.578& 
0.500&
0.600\\
new-thyroid1& 
0.962& 
0.894& 
0.900&
0.983\\
new-thyroid2& 
0.940& 
1.000& 
0.875&
0.957\\
vehicle3& 
0.694& 
0.694& 
0.536&
0.696\\
winequality-red-4& 
0.521& 
0.500& 
0.524&
0.712\\
wisconsin4& 
0.968& 
0.973& 
0.964&
0.977\\
yeast-0-2-5-6-vs 3-7-8-9& 
0.691& 
0.584& 
0.623&
0.835\\
yeast1& 
0.650& 
0.608& 
0.500&
0.682\\
glass0& 
0.808& 
0.728& 
0.632&
0.735\\
glass6& 
0.961& 
0.900& 
0.792&
0.992\\
pima& 
0.688& 
0.704& 
0.628&
0.729\\
africa recession& 
0.574& 
0.545& 
0.552&
0.586\\
insurance& 
0.654& 
0.500& 
0.500&
0.812\\
\hline
Avg.AUC&
0.759&
0.737&
0.667&
0.828\\
Avg.Rank&
2.333&
2.833&
2.722&
1.111\\
Win/Tie/Loss&
17/0/1&
17/0/1&
18/0/0&
Base\\
Fr.T&
\checkmark&
\checkmark&
\checkmark&
Base\\
\hline
\end{tabular}
\end{center}
\end{table}

\section{Conclusion and future work}
\quad This paper comes up with a new decision forest building method MA-H-SAC-DF, which targets maximizing long-term returns via deep reinforcement learning and is different from bagging and boosting. In MA-H-SAC, we firstly model the building process as De-POMDP, and all base classifiers are constructed by a set of cooperative agents jointly. Besides, global state and local observations are defined based on parent node information and location information, while the action space and reward function are the same as our previous work. Last, the Hybrid SAC is extended to MAS under the CTDE framework to find an optimal decision forest building policy. The experiment results indicate that MA-H-SAC-DF has better performance on imbalanced data. 

In the future, we will continue to explore the efficiency optimization and scalability of MA-H-SAC-DF and extend it to discrete attribute problems. Another area we would like to explore is the extension of MA-H-SAC-DF under multi-class classification scenarios.

\end{document}